\documentclass{bmvc2k}

\title{Learning Object Location Predictors with Boosting and Grammar-Guided
  Feature Extraction}

\addauthor{Damian Eads}{http://www.cs.ucsc.edu/~eads}{1} 
\addauthor{Edward Rosten}{http://mi.eng.cam.ac.uk/~er258}{2} 
\addauthor{David Helmbold}{http://www.cs.ucsc.edu/~dph}{1} 

\addinstitution{
 Department of Computer Science\\
 University of California\\
 Santa Cruz, California, USA
}
\addinstitution{
 Department of Engineering\\
 University of Cambridge\\
 Cambridge, UK
}

\usepackage{comment}
\usepackage{color}
\usepackage{amssymb,amsmath}
\usepackage{xspace}
\usepackage{tabularx}
\newcommand{\dprod}[1]{ \mathtt{#1} }

\newcommand{\dfunc}[1]{ \mbox{#1} }

\newcommand{\drand}[1]{ #1 }

\newcommand{\DEnote}[1]{}

\newcommand{\Beamer}{{\sc Beamer}\xspace}

\runninghead{Eads, \bmvaEtAl}{Location-based, Grammar-guided Object Detection}

\def\etal{\emph{et al}\bmvaOneDot}

\begin{document}
\maketitle
\begin{abstract}

We present \Beamer: a new spatially exploitative approach to learning
object detectors which shows excellent results when applied to the
task of detecting objects in greyscale aerial imagery in the presence
of ambiguous and noisy data. There are four main contributions used to
produce these results. First, we introduce a grammar-guided feature
extraction system, enabling the exploration of a richer feature space
while constraining the features to a useful subset. This is specified
with a rule-based generative grammar crafted by a human
expert. Second, we learn a classifier on this data using a newly
proposed variant of AdaBoost which takes into account the spatially
correlated nature of the data. Third, we perform another round of
training to optimize the method of converting the pixel
classifications generated by boosting into a high quality set of
$(x,y)$ locations. Lastly, we carefully define three common problems in
object detection and define two evaluation criteria that are tightly
matched to these problems. Major strengths of this approach are: (1) a
way of randomly searching a broad feature space, (2) its performance
when evaluated on well-matched evaluation criteria, and (3) its use of
the {\em location} prediction domain to learn object detectors as well
as to generate detections that perform well on several tasks:
object counting, tracking, and target detection. We demonstrate the
efficacy of \Beamer with a comprehensive experimental evaluation on a
challenging data set.

\end{abstract}

\section{Introduction}
\begin{figure}
\begin{center}
\begin{tabular}{cccp{1.1cm}p{1.1cm}p{1.1cm}}
\includegraphics[height=3.7cm]{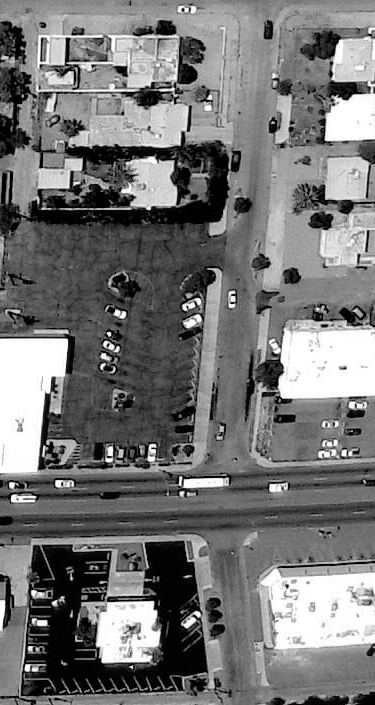}&
\includegraphics[height=3.7cm]{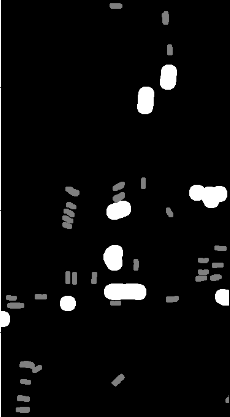}&
\includegraphics[height=3.7cm]{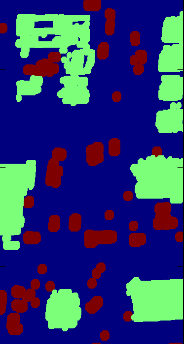}&

\begin{tabularx}{4cm}[b]{ccc}
\includegraphics[height=1.5cm]{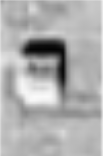} &
\includegraphics[height=1.5cm]{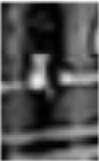} &
\includegraphics[height=1.5cm]{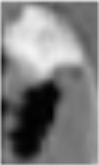} \\
(d) & (e) & (f) \\ 
\includegraphics[height=1.5cm]{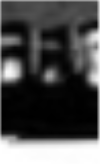} &
\includegraphics[height=1.5cm]{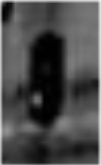} &
\includegraphics[height=1.5cm]{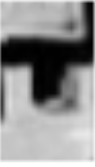}
\end{tabularx} \\
(a)&(b)&(c)&\centering\rule{.90em}{.0ex}(g)&\centering(h)&\centering\rule{-1.1em}{.0ex}(i)
\end{tabular}
\end{center}
\caption{An aerial photo of Phoenix, AZ was divided into 11 slices. An
  example slice is shown in subfigure (a). Its mark-up is shown in
  subfigure (b); background pixels are black, object pixels are
  grey, and confuser pixels, white. Subfigure (c) shows an example of
  a post processing applied to a weak hypothesis, which helps
  disambiguate between similar car and building patches by
  abstaining on building pixels. Cars are indicated by red, background by blue
  and abstention by green.
  Examples of ambiguous objects include (d) a roof-mounted
  air-conditioner, (e) an overhead street sign, (f) vegetation, (g)
  closely packed cars, (h) a dark car, and (i) a car on a roof carpark
  in partial shadow. \label{fig:hard}}
\label{fig:teaser}
\end{figure}

Learning to detect objects is a subfield of computer vision that is
broad and useful with many applications. This paper is concerned with
the task of {\em unstructured object detection}: the input to the
object detector is an image with an unknown number of objects present,
and the output is the locations of the objects found in the form of
$(x,y)$ pairs, and perhaps delineating them as well. A typical
application is detection of cars in aerial imagery for purposes such
as car counting for traffic analysis, tracking, or target
detection. Figure~\ref{fig:teaser} shows (a) an example image from the
data set used in the experiments, (b) its mark-up, (c) an example of
an initial confidence-rated weak hypothesis learned on it, and
subfigures (d-i) show some of the trickier examples in the data set.

Section~\ref{sec:background} reviews common approaches to object
detection. Section~\ref{sec:segment} describes a new variant of
AdaBoost that takes into account the spatially correlated nature of
the data to reduce the effects of label noise, simplify solutions, and
achieve good accuracy with fewer
features. Section~\ref{sec:featureextraction} describes our technique
for generating features randomly but guided by a stochastic grammar
crafted by a domain expert to make useful features more likely, and
unhelpful features, less likely. A second round of training involves
learning detectors which predict $(x,y)$ locations of objects from
pixel classifications, described in Section~\ref{sec:locations}. Since
the quality of detections greatly depends on the problem at hand, two
different evaluation criteria are carefully formulated to closely
match three common problems: tracking, target detection, and object
counting. Lastly, in our evaluation Section~\ref{sec:evaluation}, each
component in the detection pipeline is isolated and compared against
alternatives through an extensive validation step involving a grid
search over many parameters on the two different metrics. The results
are used to gain insights into what leads to a good object
detector. We have found our contributions give better results.

\section{Background}
\label{sec:background}

Localizing objects in an image is a prevalent problem in computer
vision known as {\em object detection}. {\em Object recognition}, on
the other hand, aims to identify the presence or absence of an object
in an image. Many object detection approaches reduce object detection
to object recognition by employing a {\em sliding
  window}~\cite{viola2001rod,ferrari08,laptev2006iod}, one of the more common design
patterns of an object detector. A fixed sized rectangular or circular
window is slid across an image, and a classifier is applied to each
window. The classifier usually generates a real-valued output representing
confidence of detection. Often this method must
carefully arbitrate between nearby detections to achieve
adequate performance. 

Object detection models can be loosely be broken
down into several different overlapping categories.  {\em Parts-based
  based} models consider the presence of parts  and (usually) 
the positioning of parts in relation to one
another~\cite{fergus03ocr,agarwal2004ldo,chum2007exemplar}. A special case is the {\em bag
  of words model}
where predictions are
made simply on the presence or absence of parts
rather than their overall structure or relative 
positions~\cite{heisele2003frc,zhang2007lfa}.  Some parts-based models
model objects by their characterizing shape during
learning and matching shape to detect~\cite{belongie02}.
{\em Cascades} are commonly
used to reduce false positives and improve computational
efficiency. Rather than applying a single computationally expensive
classifier to each window, a sequence of cheaper classifiers is
used. Later classifiers are invoked only if the previous classifiers
generate detections. {\em Generative model} approaches learn a
distribution on object appearances or
object configurations~\cite{mikolajczyk2006moc}. 
{\em Segmentation-based} approaches fully
delineate objects of interest with polygons or pixel
classification~\cite{shotton2008stf}. {\em Contour-based} approaches
identify contours in an image before generating
detections~\cite{ferrari08,opelt2006bfm}. {\em Descriptor vector}
approaches generate a set of features on local image patches. One of
the most commonly used descriptors is the Scale Invariant Feature
Transform (SIFT), which is invariant to rotation, scaling, and
translation and robust to illumination and affine
transformations~\cite{lowe99sift}. 
A large
number of object detectors use {\em interest point detectors} to find
salient, repeatable, and discriminative points in the image as a first
step~\cite{dorko03,agarwal2004ldo}. Feature descriptor vectors are
often computed from these interest points. {\em Probabilistic models}
estimate the probability of an object of interest occurring; generative
models are often used~\cite{schneiderman00,fergus03ocr}. 
{\em Feature Extraction} creates higher level 
representations of the image that are often easier for algorithms to learn
from. Heisele, \etal~\cite{heisele2003frc} train a two-level hierarchy
of support vector machines: the first level of SVMs finds the presence
of parts, and these outputs are fed into a master SVM to determine the
presence of an object. Dorko, \etal~\cite{dorko03} use an interest
point detector, generate a SIFT description vector on the interest
points, and then use an SVM to predict the presence or absence of objects.

One of the more popular and highly regarded feature-based object
detectors is the sliding window detector proposed by Viola and
Jones~\cite{viola2001rod}, which uses a feature set originally
proposed by Papageorgiou, \etal~\cite{papa98}. 
Adjacent rectangles of equal size are filled with $1$s and $-1$s and
embedded in a kernel filled with zeros. The kernel is convolved
with the image to produce the feature and using an
integral image greatly reduces the computation time for these
features.
Viola and Jones employ a cascaded sliding window approach where each
component classifier of the cascade is a linear combination of weak
classifiers trained with AdaBoost. 


\section{Approach}
\begin{figure}
\begin{center}
\includegraphics[width=5in]{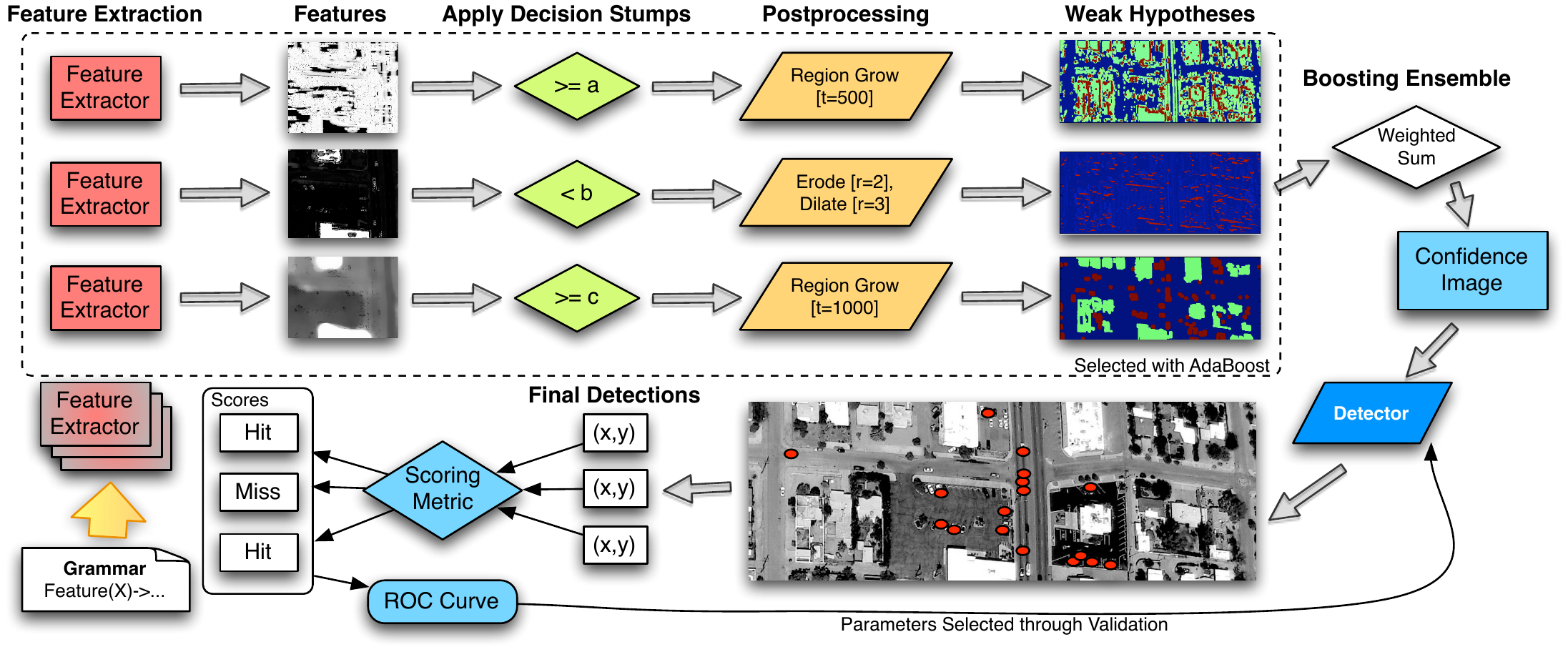}
\end{center}
\caption{
Object detection is carried out in a pipeline consisting of three
stages: feature extraction, pixel classification, and locality
predictions in the form of $(x,y)$. At each training iteration, a new
pool of feature extractors generated by a grammar. \Beamer then
chooses the best feature extractor, decision stump, and
post-processing filter combination. Thresholding these features yields
a weak pixel classification which are combined with AdaBoost to
produce a confidence image. The grey arrows show the flow of data to
carry out object detection from start to finish for a static instance
of an object detector.}
\label{fig:system}
\end{figure}

The \Beamer object detector pipeline consists of a feature extraction
stage, pixel classification stage, and a detector stage as
Figure~\ref{fig:system} shows.  First, a set of learned features are
combined into a pixel classifier using
AdaBoost~\cite{freund96experiments}.  Then, the detector
pipeline (see Section~\ref{sec:locations}) transforms the pixel
classifications into a set of $(x,y)$ locations representing the
predicted locations of the objects.  Our methodology partitions the
data set into training, validation, and test image sets. The pixel
classifier is learned during the training phase on the training images
with the grammar constraints, post-processing parameters, and stopping
conditions remaining fixed. These fixed parameters are later tuned on
the validation set along with the detector's parameters. The detector
generates $(x,y)$ location predictions from the pixel
classification. After the training and validation steps, a fully
learned object detector results. The grey arrows show how data flows
through a specific instance of an object detector.

Section~\ref{sec:featureextraction} describes the very first step of
weak pixel classification, feature extraction, which is carried out by
generating features with a generative grammar. Section~\ref{sec:segment}
describes the learning of an ensemble of weak pixel classifications
using boosting. Finally, Section~\ref{sec:locations} explains how the
pixel classification ensemble is transformed into $(x,y)$ location
domain predictions. A complete list of all the parameters described 
in the following sections is given in Table~\ref{tbl:params}. 

\subsection{Feature Extraction}
\label{sec:featureextraction}
A single pixel in a greyscale image provides very limited information
about its class. Feature extraction is helpful for generating a more
informative feature vector for each pixel, ideally incorporating
spatial, shape, and textural information. This paper considers
extracting features with {\em neighborhood} image operators such as
convolution and morphology. Even good sets of neighborhood-based
features are unlikely to have enough information to perfectly predict
labels, but the hope is that large and diverse sets of features can
encode enough information to make adequate predictions. At each
boosting iteration, a new set of random features is generated, but
only the best feature of this set is kept.

\begin{figure}
\begin{center}
\scalebox{0.8}{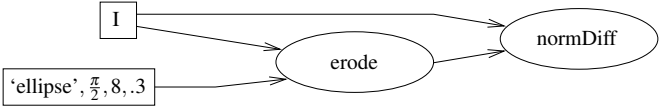}
\end{center}
\caption{ An example of a feature extractor program generated by the
  \Beamer grammar which is achieved by reducing $\dprod{Feature}(I)$
  using the production rules of the grammar (where $I$ is an image
  variable), {\small $\dprod{Feature}(I)
    \rightarrow \dprod{Compound}(I)
    \rightarrow \dprod{Binary}(I, \dprod{Compound}(I))
    \rightarrow \dprod{Binary}(I, \dprod{Unary}(I))
    \rightarrow \dprod{Binary}(I, \dprod{NLUnary}(I))
    \rightarrow \dprod{Binary}(I, \dprod{Morph}(I, \dprod{RandomSE}))
    \rightarrow \dprod{Binary}(I, \dfunc{erode}(I, \dprod{RandomSE}))
    \rightarrow \dprod{Binary}(I, \dfunc{erode}(I, ('ellipse', pi/2, 8, 0.3)))
    \rightarrow \dfunc{normDiff}(I, \dfunc{erode}(I, ('ellipse', pi/2, 8, 0.3)))$}}

\label{fig:gexs}
\end{figure}

Generative grammars are common structures used in Computer Science to
specify rules to define a set of
strings~\cite{sipser2005itc,hopcroft1990iat}. Extending our earlier
work on time series~\cite{eads05tech}, we use them to specify the
space of feature extraction programs, which are represented as
directed graphs (a graph representation is preferable because it
allows for re-use of sub-computations). A grammar is made up of
nonterminal productions such as $P \rightarrow A | B$, which are
expanded to generate a new string.  The rules associated with the production
are selected at random, so $P$ can be expanded as either $A$ or $B$.
Figure~\ref{fig:gexs} shows an
example graph program generated by the \Beamer~grammar shown in
Figure~\ref{fig:grammar}.  The primitive operators used
for our object detection system are listed in Table~\ref{tbl:ops} and
the grammar governing how they are combined is shown in
Figure~\ref{fig:grammar}.

\graphicspath{{.}{images/}}

\begin{small}
\begin{table}
\begin{center}
\begin{tabular}{|l|p{3.4in}|}
\hline
\bf Function & \bf Description \\
\hline\hline
mult($I_A$, $I_B$) & Element-wise multiplies two images, $f(a,b)=ab$.\\
\hline
blend($I_A$, $I_B$) & Element-wise averaging of two images, $f(a,b)=\frac{a+b}{2}$.\\
\hline
normDiff($I_A$, $I_B$) & Normalized difference, $f(a,b)=\frac{a-b}{\sum_{p \in I_A}{p}+\sum_{p \in I_B}{p}}$.\\
\hline
scaledSub($I_A$, $I_B$) & Scaled difference, $f(a,b)=\frac{a-b}{a+b}$.\\
\hline
sigmoid($I$, $\theta$, $\lambda$)  & Soft maximum with threshold $\theta$ and scale $\lambda$, $f(u)=\frac{\arctan(\lambda(u + \theta))}{\lambda}$ \\
\hline
ggm(I, $\sigma$) & Applies a Gaussian Gradient Magnitude to an image.\\
\hline
laplace(I, $\sigma$) & Laplace operator with Gaussian 2nd derivatives \& standard deviation $\sigma$. \\
\hline
laws(I, $u$, $v$) & Applies the Laws texture energy kernel $u \cdot v$. \\
\hline
gabor(I, $\theta$, $k$, $r$, $\nu$, $f$)  & Applies a gabor filter of a specified angle $\theta$,
            size $k$, ratio $r$, frequency $\nu$, and envelope $f$.\\
\hline
ptile($I$, $p$, S)           & A $p$'th percentile
                               filter with a structuring element $S$ applied to an image $I$.\\
\hline
\end{tabular}
\end{center}
\caption{Primitive operators used by the grammar. Element-wise operators are described by a function $f(a,b)$ of two pixels $a$ and $b$. Unary operators $f(u)$ are described by a function of one pixel $u$. A $k$ by $k$ structuring element is parametrized with an ellipse orientation $\theta$ and width to height ratio $r$.
}
\label{tbl:ops}
\end{table}
\end{small}

\def\Alternate{~~~|}
\begin{figure}
\begin{center}
\scalebox{0.8}{\parbox{1.2\textwidth}{
\begin{eqnarray*}
\dprod{Feature}(X)     & \rightarrow & \dprod{Binary}(\dprod{Unary}(X), \dprod{Unary}(X))\\
                       & \Alternate  & \dprod{NLUnary}(\dprod{Unary}(X))\\
                       & \Alternate  & \dprod{NLBinary}(\dprod{Unary}(X), \dprod{Unary}(X))\\
                       & \Alternate  & \dprod{Compound}(X)\\
\dprod{Binary}(X, Y)   & \rightarrow & \dfunc{mult}(X, Y)~|~\dfunc{normDiff}(X, Y)~|~\dfunc{scaledSub}(X, Y) ~|~ \dfunc{blend}(X, Y)\\
\dprod{NLBinary}(X, Y)   & \rightarrow & \dfunc{mult}(X, Y)~|~\dfunc{normDiff}(X, Y)\\
\dprod{Unary}(X)       & \rightarrow & \dprod{LUnary}(X) ~|~ \dprod{NLUnary}(X) \\
\dprod{Compound}(X)    & \rightarrow & \dprod{Unary}(X) ~|~ \dprod{Binary}(X, \dprod{Compound}(X)) \\
\dprod{Morph}(X, S)    & \rightarrow & \dfunc{erode}(X, S) ~|~ \dfunc{dilate}(X, S) ~|~ \dfunc{open}(X, S) ~|~ \dfunc{close}(X, S)\\
\dprod{RandomSE}()       & \rightarrow & (\drand{\theta} \in [0, 2\pi], \{2k + 1 | \drand{k} \in \{1, \ldots, 7\}\}, \{10^{2s-1} | s \in [0, 1]\})\\
\dprod{NLUnary}(X)     & \rightarrow & \dfunc{sigmoid}(X, \drand{a} \in \mbox{SNorm}(), \drand{b} \in \{0.1, 0\})\\
                       & \Alternate  & \dprod{Morph}(X, \dprod{RandomSE}()) \\
                       & \Alternate  & \dfunc{ptile}(X, p \in [0, 100], \dprod{RandomSE}())\\
                       & \Alternate  & \dfunc{ggm}(X, 3 * \mbox{SNorm}()) \\
\dprod{LUnary}(X)      & \rightarrow & \dfunc{laws}(X, u \in \{L_5,E_5,S_5,R_5,W_5\}, v \in \{L_5,E_5,S_5,R_5,W_5\})\\
                       & \Alternate  & \dfunc{laplace}(X, \drand{\sigma} \in 3*\mbox{SNorm}())\\
                       & \Alternate  & \dfunc{gabor}(X, \drand{\theta} \in [0, \pi], \drand{k} \in [1, 31],\{10^{2q-1} | \drand{q} \in [0, 1]\}, \{10s+2 | s \in [0, 1]\}, \dfunc{sin}|\dfunc{cos}|\dfunc{both})\\
                       & \Alternate  & \dfunc{convolve}(X, \dprod{ViolaJonesKernel}())\\
\end{eqnarray*}}}
\end{center}
\caption{The grammar used to generate features for the pixel
  classification stage of the object detection system. The
  $\dprod{ViolaJonesKernel}()$ does not sample uniformly from the space of all
  kernels. Rather, the kernel type (horizontal-2, vertical-2, horizontal-3, vertical-3, quad) is
  chosen uniformly at random, followed by the size, then location. $\dprod{RandomSE}$ defines an elliptical
  structuring element, where the parameters of the ellipse are respectively orientation, major radius in pixels
  and aspect ratio. The meanings of the other parameters are given in Table~\ref{tbl:ops}.
  }
\label{fig:grammar}
\end{figure}

\begin{figure}
\begin{center}
\begin{tabular}{ccccc}
\includegraphics[width=2.2cm]{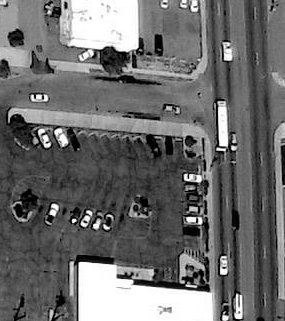} &
\includegraphics[width=2.2cm]{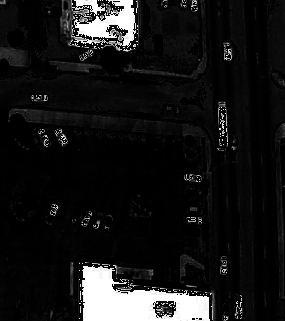} &
\includegraphics[width=2.2cm]{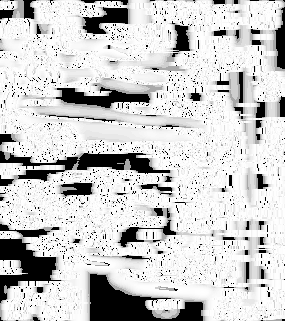} &
\includegraphics[width=2.2cm]{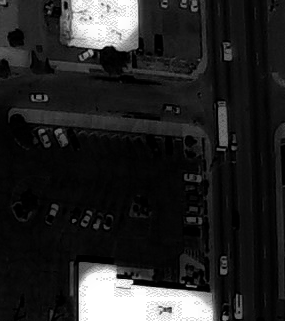} &
\includegraphics[width=2.2cm]{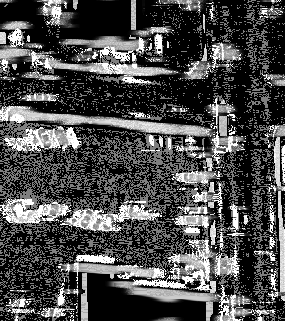}\\
(a)&(b)&(c)&(d)&(e)\\
\end{tabular}
\end{center}
\caption{Subfigures (b)-(e) are five examples of features generated by a grammar and applied to the image shown in subfigure (a).}
\label{fig:exs}
\end{figure}

\subsection{Pixel Classification with Spatially Exploitative AdaBoost}
\label{sec:segment}

The top of Figure~\ref{fig:system} illustrates the pixel
classification part of the \Beamer~object detection pipeline. The goal
of pixel classification is to fully delineate the class of interest but we introduce modifications.
A set of feature extraction algorithms is applied to an image,
resulting in a set of feature images.  These feature images are
thresholded and post-processed to create weak pixel classifiers for
detecting object pixels.  The final pixel classifier is a weighted
combination of these weak pixel classifiers which output confidence
with their predictions.

Learning is based on a training set 
where all the pixels belonging to the objects of interest (cars in our case)
are hand-labeled.  
There are several difficulties in identifying good weak pixel classifiers from
the hand-labeled training data.
First, in applications like ours there are many more  background
pixels than foreground (object) pixels.
Providing too much background puts too much
emphasis on the background during learning, and can lead to hypotheses
that do not perform well on the foreground.
Second, hand-labeling is a subjective and error prone activity. 
Pixels outside
the border of the object may be accidentally labeled as car, and
pixels inside the border as background. It is well known that label
noise causes difficulties for AdaBoost~\cite{dietterich2000aec, ratsch2001sma}.
This difficulty is compounded when the image data itself
is noisy or there may not be sufficient information in a pixel
neighborhood to correctly classify every pixel. 
Third, training a
pixel classifier that fully segments is a much harder problem than
localization. For example, if a weak hypothesis correctly labels 
only a tenth of the object pixels and these correct
predictions are evenly distributed throughout the objects, the weak
hypothesis will appear unfavorable.
This is unfortunate because the weak hypothesis may be very good at
localizing objects, just not fully segmenting them. 
Similarly, some otherwise good 
features may identify many objects as well as large swaths of background. In
terms of localization, the performance is good
but these hypotheses will be rejected by the
learning algorithm because of the large number of false positives they
produce.

We propose three spatially-motivated modifications to standard
AdaBoost to perform well with the difficulties above.  First, we weight the
initial distribution so the sum of the foreground weight is
proportionate with the background class.  Second, we use
confidence-rated AdaBoost proposed by Schapire and
Singer~\cite{schapire1999iba} so weak hypotheses can output low or
zero confidence on pixels which may be noisy or labeled incorrectly.
In confidence-rated boosting, the weak hypotheses output predictions
from the real interval $[-1,1]$ and the more confident predictions are
farther from zero. In the boosting literature, the {\em edge} is
defined as the weighted training error.
Third, we perform post-processing on the weak pixel classifications to improve those
that produce good partial segmentations of objects. 

\paragraph{Weak Classifier Post-processing} Four different weak pixel classification post-processing filters are
considered and compared against no filtering at all. The first
technique performs {\bf region growing} (abbreviated R) with a
4-connected flood fill. Regions larger than $k$ pixels are identified
and converted to abstentions (zero confidence predictions). This is
useful for disambiguating cars from large swaths, such as buildings,
which may have similar texture as cars. This simple post-processing
filter works very well in practice. The other three post-processing
techniques apply either an {\bf erosion} (E), {\bf a dilation} (D), or
a local {\bf median filter} (M) using a circular structuring element of
radius $r$. When applying one of these filters, a pixel classifier
only partially labeling an object will be evaluated more
favorably. This improves the stability of learning in situations where
the object pixels are noisy in the images and pixels are
mislabeled. Section~\ref{sec:evaluation} thoroughly compares the
performance of different combinations of these four post-processing
filters, all of which show better performance than no filtering at
all.

\subsection{Learning and Predicting in the $(x,y)$ Location Domain}
\label{sec:locations}
The final stage of object detection turns the confidence-rated pixel
classification into a list of locations pointing to the objects in an
image. Noisy and ambiguous data often reduce the quality of the pixel
classification, but since we use pixel classification as a step along
the way, we perform an extra round of training to learn to transform a
rough labeling of object pixels into a high quality list of locality
predictions, and to do so in a noise-robust and spatially exploitative
manner. Pure pixel-based approaches are hard to optimize for location-based
criteria, and often translate mislabeled pixels into false
positives. Our algorithms turn the pixel classifications
into a list of object locations, allowing us to operate in and directly optimize over
the same domain as the output: a list of $(x,y)$ locations. 

A confidence-rated pixel classification provides predictive power
about which pixels are likely to belong to an object. The goal is a
high quality localization, rather than object delineation, so we reduce the set of positive pixels to a
smaller set of high-quality locations. The first object detector, {\bf Connected
Components (CC)} thresholds the confidence image at zero, performs binary
dilation with a circular structuring element of radius $\sigma_{CC}$, 
finds connected components and marks
detections at the centroids of the components.

{\bf
  Large Local Maxima (LLM)}, is like non-maximal suppression but
instead represents the locations and magnitudes of the maxima in location
space as opposed to image space. The approach sparsifies the set of
high confidence pixels by including only local maxima as guesses of an
object's location. Next, the LLM detector chooses among the set of
local maxima those pixel locations with confidences exceeding a
threshold $\theta_{LLM}$. This method of detection is attractive
because it is very fast, and somewhat reminiscent of decision
stumps. The detector outputs these large maxima as its final predictions,
ordering them with decreasing confidence. A Gaussian smoothing of
width $\sigma_{LLM}$ can be applied before finding the maxima 
to reduce the noise and further
refine the solutions.

The LLM detector treats maxima locations independently, which can be
quite sensitive to the presence of outlier pixels and noisy imagery.
Noisy imagery often leads to an excess of local maxima, some of which
lie outside an object's boundary, which often results in false
positives. We propose an extension of the LLM detector called the {\bf
  Kernel Density Estimate (KDE)} detector for combining maxima
locations into a smaller, higher quality set of locations based on
large numbers of maxima with high confidence clustered spatially close
to one another. More specifically, the final detections are the modes
of a confidence-weighted Kernel Density Estimate computed over the set
of LLM locations. The width of the kernel is denoted
$\sigma_{KDE}$. Our results show the KDE and LLM detectors perform
remarkably well in the presence of noise.

\section{Evaluation and Conclusions}
\label{sec:evaluation}

\begin{figure}
\begin{center}
\begin{tabular}{ccc}
\includegraphics[width=4.0cm]{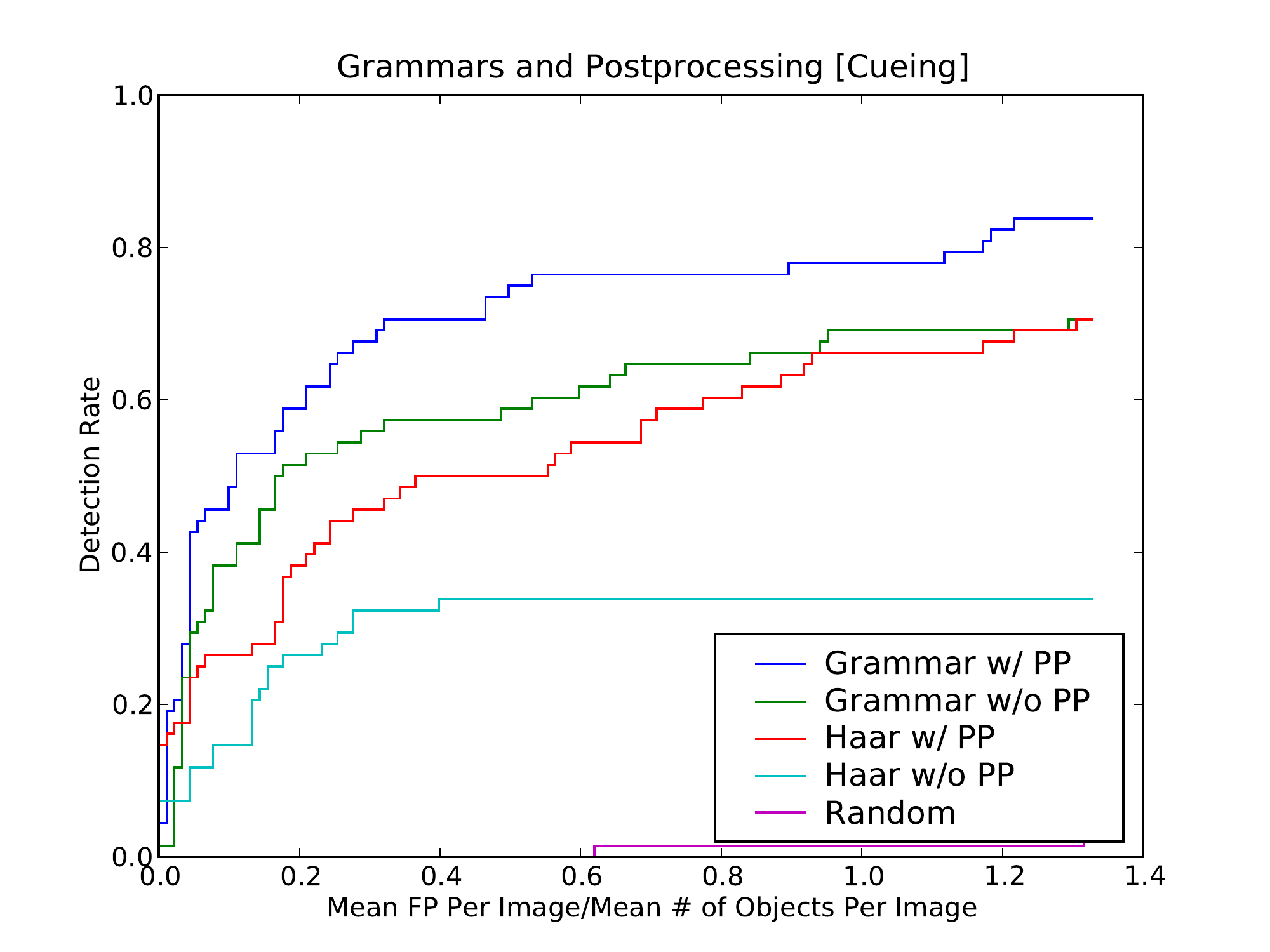} &
\includegraphics[width=4.0cm]{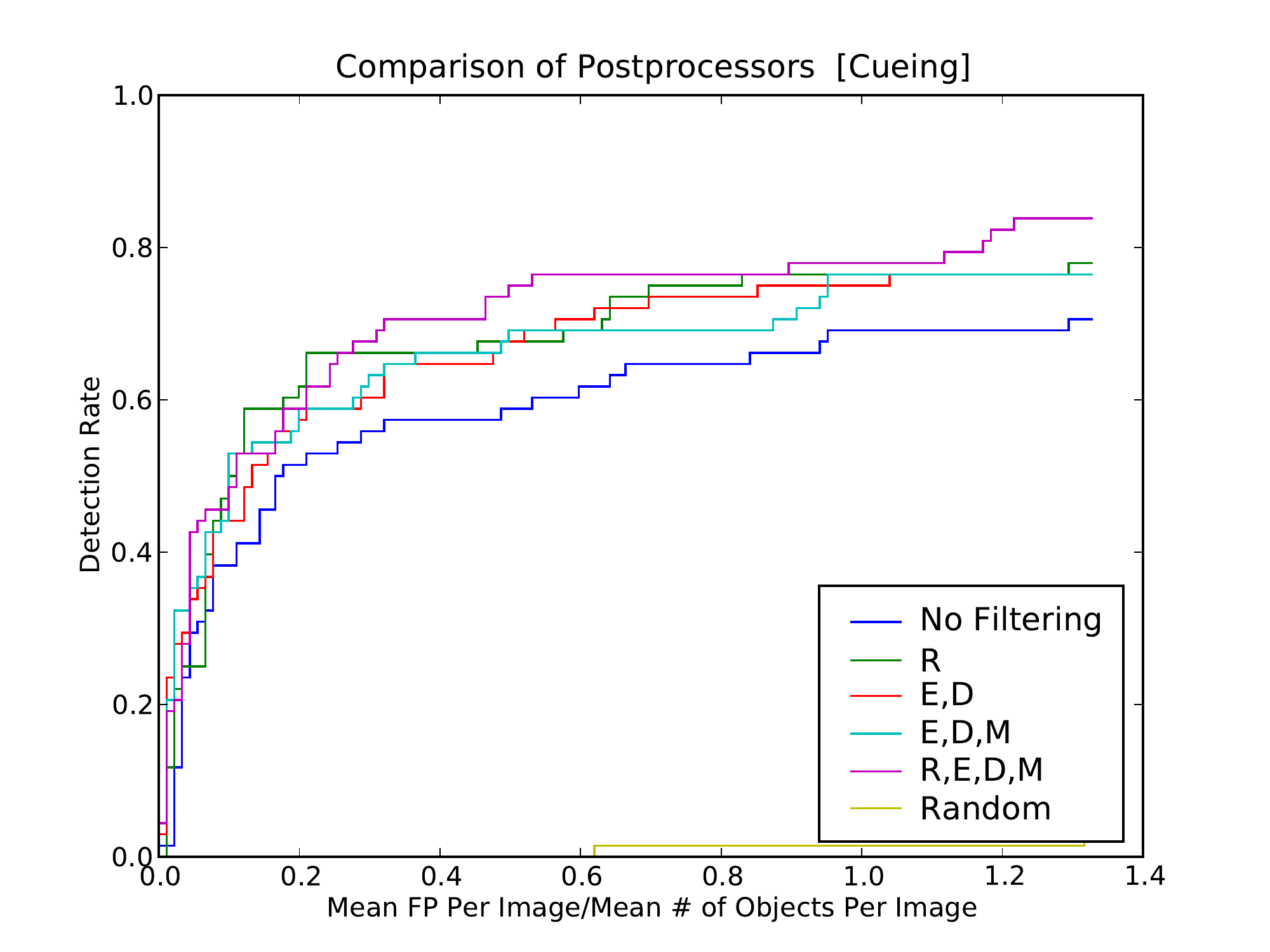} &
\includegraphics[width=4.0cm]{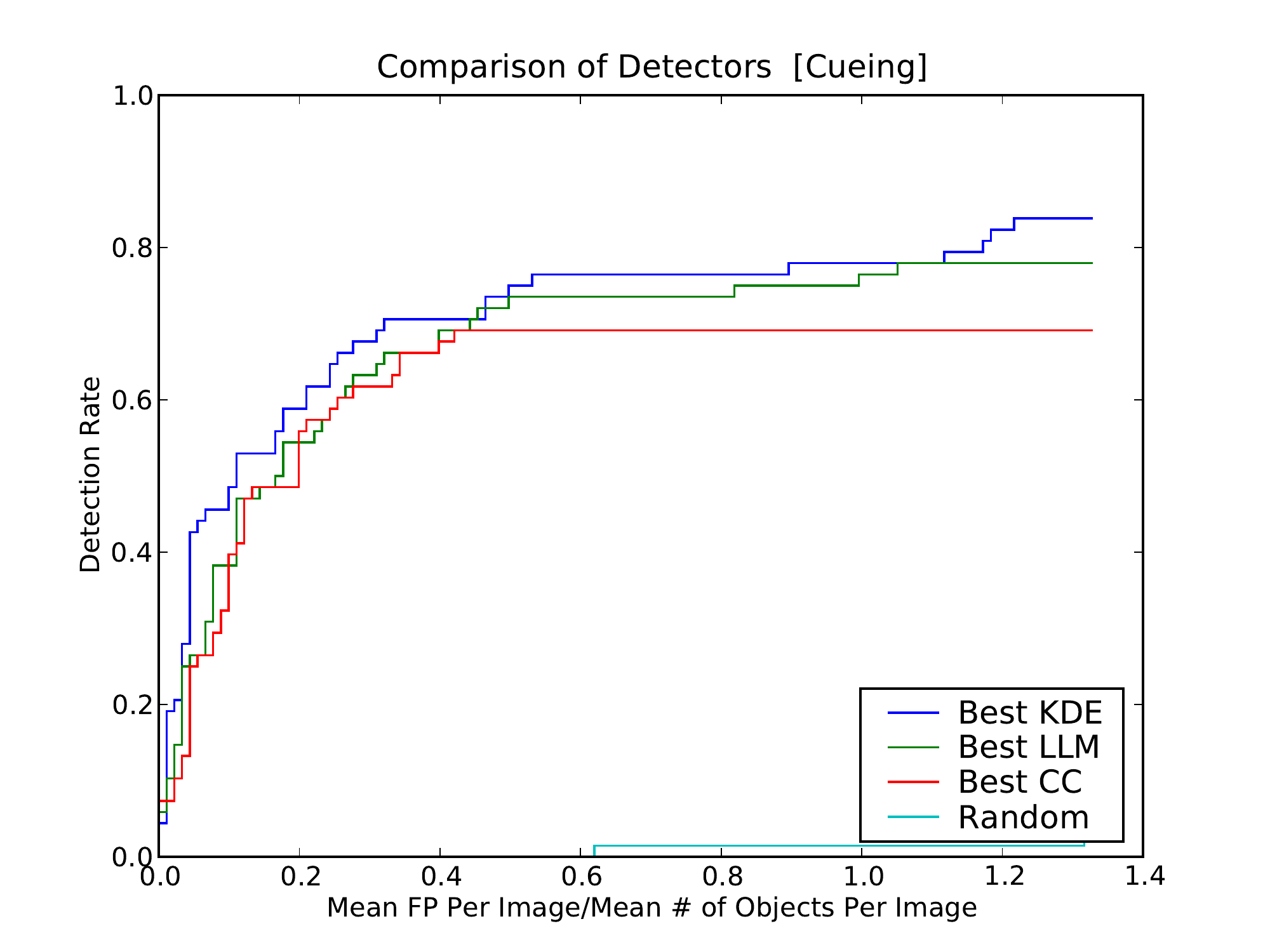} \\
(a)&(b)&(c)\\
\includegraphics[width=4.0cm]{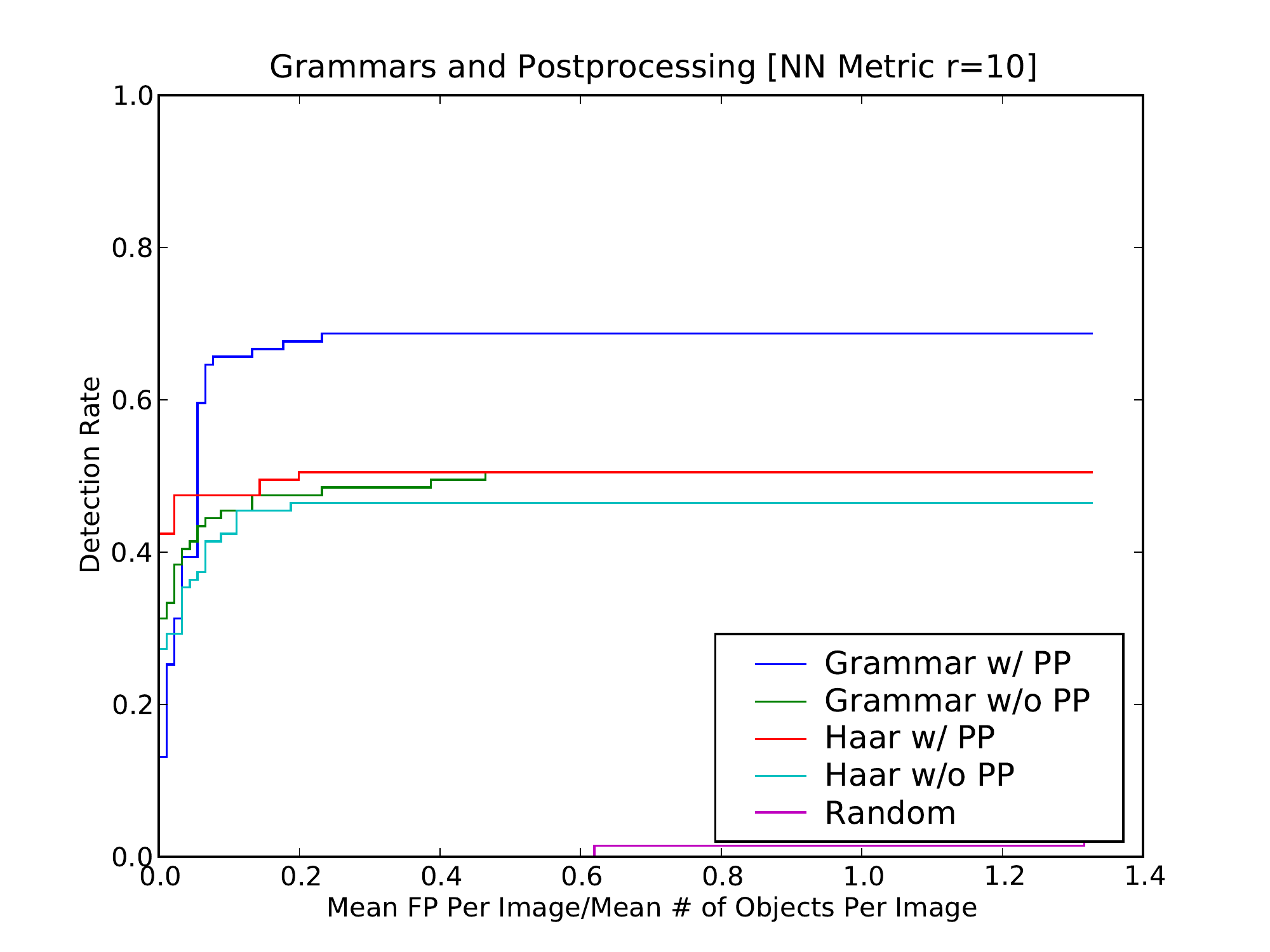} &
\includegraphics[width=4.0cm]{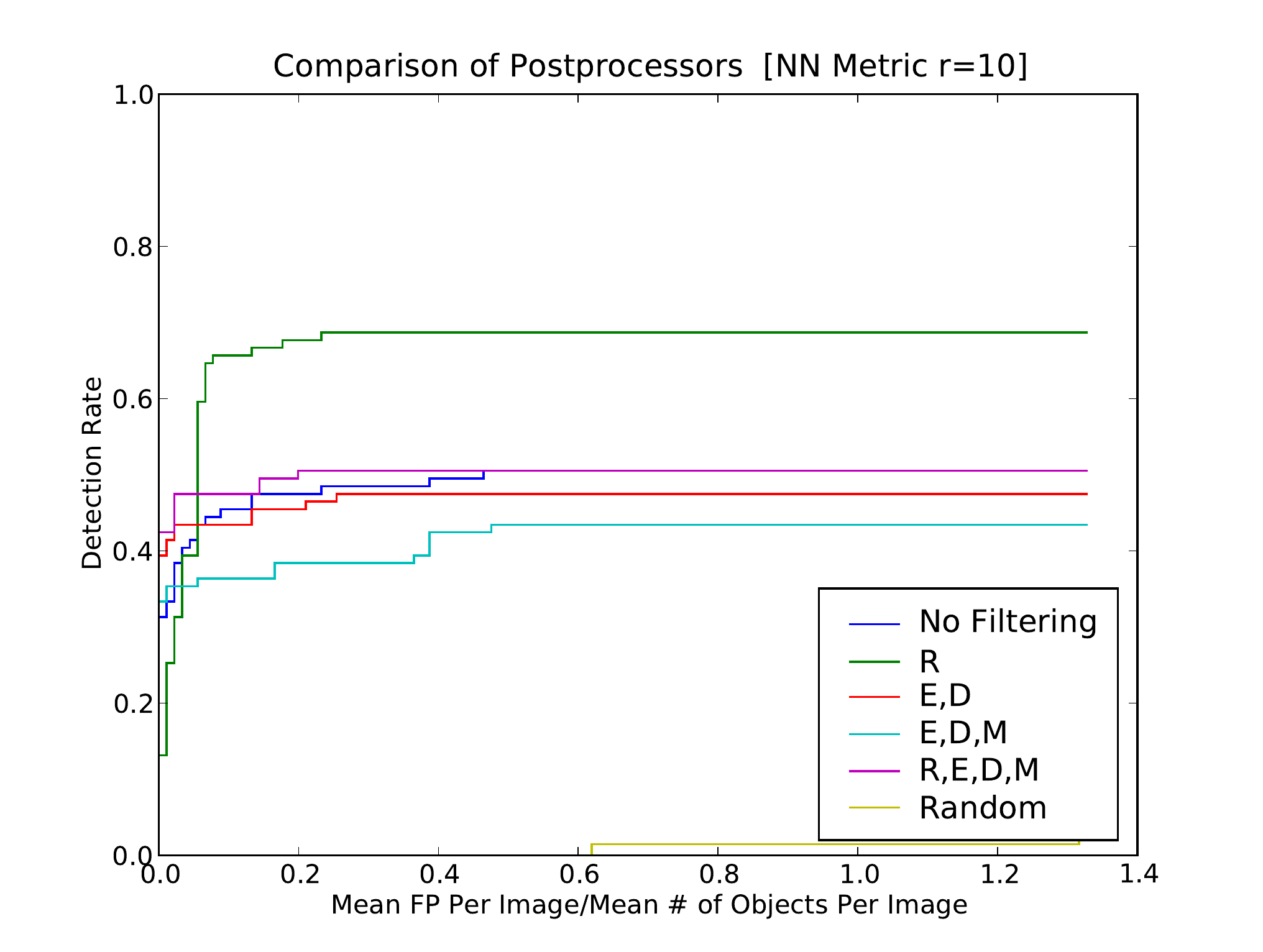} &
\includegraphics[width=4.0cm]{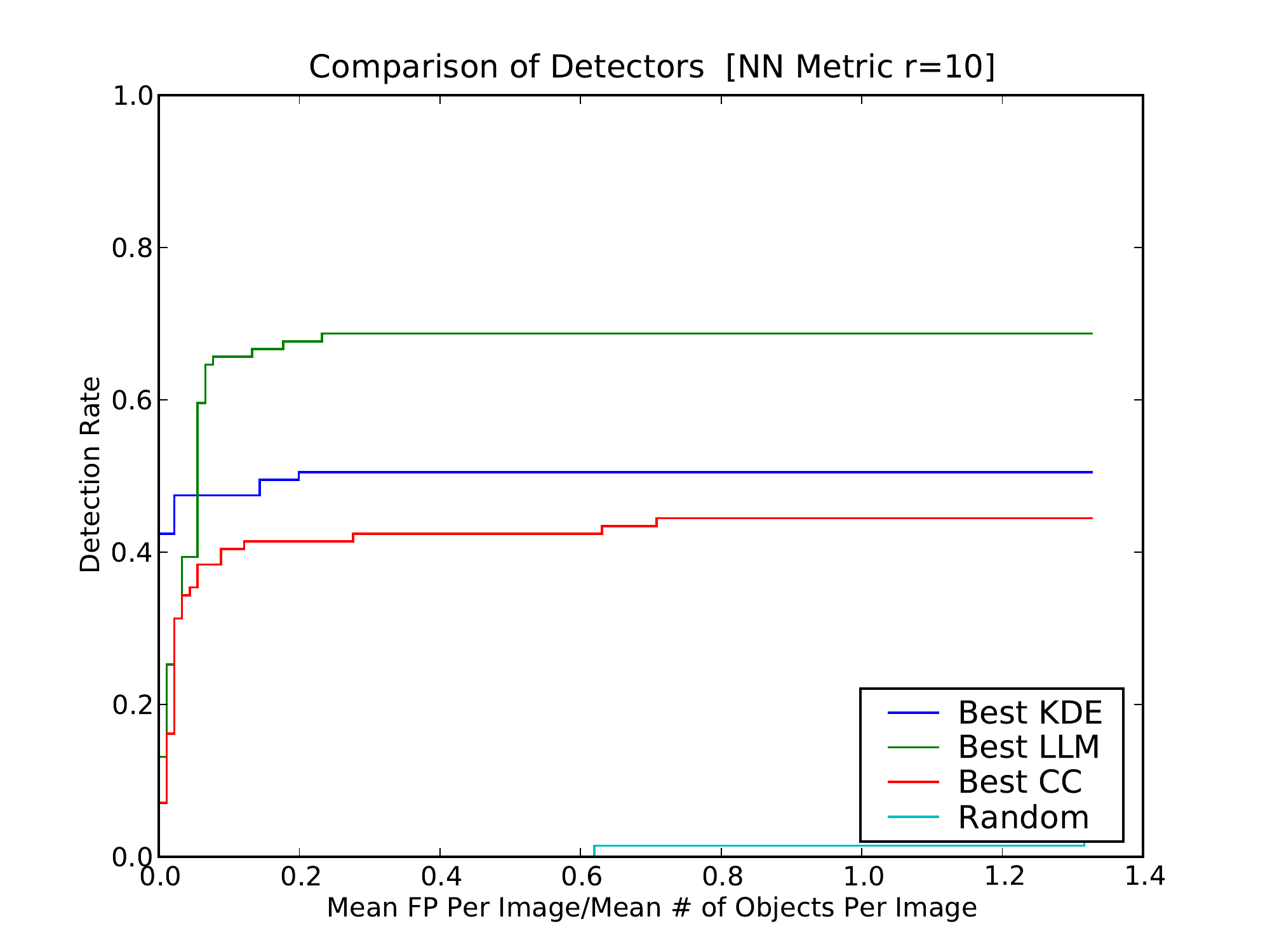} \\
(d)&(e)&(f)\\
\end{tabular}
\end{center}
\caption{Subfigures (a-c) show the result of applying the best model
  to the validation set using the cueing metric. Subfigures (d-f) show
  the result for the nearest neighbors metric. The best model for each
  aspect in a comparison filter is applied to the unseen test data
  set.}
\label{fig:roc}
\end{figure}

Generating a ROC curve for a classifier involves marking each
classification as a true negative or false positive. Quantifying the
accuracy of unstructured object detection with a ROC curve is not as
straightforward: the criteria for marking a {\em true positive} or
{\em false positive} depends on the object detection task at hand. We
consider three object detection problems: {\bf cueing}, {\bf
  tracking}, and {\bf counting} and define two criteria to mark
detections that are closely matched with these problems. Points on the ROC
curves are then drawn using predicted locations above some confidence threshold.

\begin{table}
\begin{center}
\begin{tabular}{|l|l|l|}
\hline
\bf Parameter        & \bf Parameters Tried                                          \\
\hline\hline
Iterations       & $T \in \{10, 25, 50, 75, 100\}$   \\
\hline
Features/per iter& $w=100$  \\
\hline
Feature Set      & Grammar, Haar-only, Grammar w/o Morphology or w/o Haar   \\
\hline
Post-Processing  & Region grow (R), Erosion (E), Dilation (D), Median (M), None (N)  \\
                 & w/ combinations $\{R\}$, $\{E,D\}$, $\{E,D,M\}$, $\{R,E,D,M\}$, $\{N\}$ \\
CC Detector      & $\sigma_{CC}$ from 0 to 20 (0.2 increments) exclusive. \\
\hline
LLM Detector     & $\sigma_{LLM}$ from 0 to 20 (0.2 increments) exclusive. \\
\hline
KDE Detector     & $\sigma_{KDE}$ from 0 to 10 (0.1 increments) exclusive, $\sigma_{LLM}$ as above. \\
\hline \hline
Features          & Generate $w$ for each of $T$ iterations. \\
\hline
Decision Stump    & Pick best threshold for each post-processing parameter tried. \\
\hline
Region grow PP    & $k$ is varied from 1000 to 5000 (increments of 500).  \\
\hline
Region grow PP    & $k$ is varied from 1000 to 5000 (increments of 500).  \\
\hline
E,D,M PP          & $r$ is varied between $1$ and $5$.\\
\hline

\hline
\end{tabular}
\end{center}
\caption{The first part of the table describes each parameter adjusted
  during validation. A highly extensive grid search was performed over
  a parameter space defined by the Cartesian product of these
  parameters. The second part shows the model parameters adjusted
  during an AdaBoost training iteration.}
  \label{tbl:params}
\end{table}

The goal of the {\bf cueing} task is to output detections within the
delineation of the object. False positives away from objects are
penalized, but multiple true positives are not. Figure~\ref{fig:roc},
subfigures (a-c) show the results for this metric. 
 
We introduce the {\bf nearest neighbors} criteria for marking
detections for {\bf object tracking}. Good detectors for tracking
localize objects within some small error, and multiple detections of a
given object are penalized. 
At each threshold the criteria finds the detection closest to an object. This
pair is removed and the process is repeated until either no detections or no
objects remain, or the distance of all remaining pairs exceeds a radius, $r$. Remaining objects are
{\em false negatives} and remaining detections are {\em false positives}.

Lastly the task of {\bf object counting} is concerned less with
localization and more with accurate counts. We employ the nearest
neighbors criteria for this purpose but to loosen the desire
for a spatial correlation between detections and object locations,
we set the nearest neighbor radius threshold $r$ to a high value. 

We use the {\bf Area Under ROC Curve} (AROC), computed numerically
with the trapezoidal rule, as the statistic to optimize during
validation to find the model parameters that perform the most
favorably on the validation set.  Since detectors may generate
vast numbers of false positives, we arbitrarily truncate the
curves at $U$ false positives per image ($U=30$ in our experiments). Validation is
performed over the range of parameters given in Table~\ref{tbl:params}.

Figure~\ref{fig:roc} illustrates the results of applying the most
favorable models and parameter vectors (determined using the validation data) to
the test set. Subfigure (a)--(c) and (d)--(f) illustrate the performance using
the cueing and tracking metrics. Subfigures (a) and (d) show the
clear advantage of using post-processing and grammar-guided
features over just Haar-like features. Subfigure (b) and (e) show the benefit of post-processing for reducing the effects of label and image noise, and clearly highlights the need to properly tune parameters through validation and train all stages for each problem. Region growing performs better on the nearest neighbors metric--unsurprising as it abstains on ambiguous background patches, reducing false positives. On the cueing metric, morphology helps in reducing the effects of label noise, which often leads to false positives outside an object delineation. Finally subfigures (c) and (f) show that the spatially exploitative detection algorithms LLM and KDE outperform the pixel-based CC detector.

\bibliography{bmvc_final}

\end{document}